# When Simpler Wins: Facebook's Prophet vs LSTM for Air Pollution Forecasting in Data-Constrained Northern Nigeria

Habeeb Balogun[a] and Yahaya Zakari[b]

[a] School of Computer Science and Engineering,
University of Westminster, London, United Kingdom
Baloguh@westminster.ac.uk

[b] Department of Statistics,
Ahmadu Bello University, Zaria, Nigeria.
yzakari@abu.edu.ng

# ABSTRACT

Air pollution forecasting is critical for proactive environmental management, yet data irregularities and scarcity remain major challenges in low-resource regions. Northern Nigeria faces high levels of air pollutants, but few studies have systematically compared the performance of advanced machine learning models under such constraints. This study evaluates Long Short-Term Memory (LSTM) networks and the Facebook Prophet model for forecasting multiple pollutants (CO, $SO_2$, $SO_4$) using monthly observational data from 2018 to 2023 across 19 states. Results show that Prophet often matches or exceeds LSTM's accuracy, particularly in series dominated by seasonal and long-term trends, while LSTM performs better in datasets with abrupt structural changes. These findings challenge the assumption that deep learning models inherently outperform simpler approaches, highlighting the importance of model-data alignment. For policymakers and practitioners in resource-constrained settings, this work supports adopting context-sensitive, computationally efficient forecasting methods over complexity for its own sake.

# 1. Introduction

Air pollution is a critical global environmental and public health challenge, responsible for an estimated 8.1 million premature deaths annually and significantly reducing life expectancy (Manisalidis et al., 2020; State of Global Air, 2023). In Nigeria, the problem is particularly severe due to rapid urbanisation, population growth, and widespread use of solid fuels. Cities like Onitsha and Kano have some of the highest global levels of particulate matter (PM10 and PM2.5), with Onitsha once ranked the world's most polluted city and Kano the most polluted in Africa (WHO, 2018). In rural and peri-urban areas, household air pollution from biomass fuels exacerbates health risks, contributing to over 64,000 deaths in Nigeria in 2017 alone (WHO, 2018).

The health impacts disproportionately affect vulnerable groups such as children, the elderly, and individuals with preexisting conditions. Fine particulate matter (PM2.5) and ozone ($O_3$) are particularly harmful, strongly linked to respiratory and cardiovascular diseases (Ojha et al., 2022). Meteorological factors: wind speed, temperature, and rainfall further influence pollutant dispersion and concentration, complicating forecasting efforts (Balogun, Alaka, & Egwim, 2021; Balogun, Alaka, Egwim, et al., 2021a, 2021b; Egwim et al., 2025).

Traditional statistical forecasting methods like ARIMA, SARIMA, and VARIMA have been used to predict air quality, but often underperform due to their assumptions of stationarity and limited capacity to capture nonlinear, multivariate dynamics (Lu et al., 2018; (Balogun, Alaka, & Egwim, 2021; Egwim et al., 2025). Consequently, machine learning (ML) and deep learning (DL) models have gained traction. Long Short-Term Memory (LSTM) networks excel at modelling long-range temporal dependencies in time series data (Zhu et al., 2021; Greff et al., 2016).

However, the effectiveness of LSTM models in environmental applications, especially in data-scarce regions like Nigeria, is uncertain. These models are sensitive to noisy, incomplete, and irregular real-world data, require extensive tuning, and risk overfitting small or erratic datasets (Lu et al., 2018; Jin et al., 2021). Their computational complexity also limits accessibility for routine use by public health or environmental agencies.

As a simpler alternative, Facebook Prophet, a time-series model designed to handle seasonality, missing data, and trend shifts, has shown promise in air pollution forecasting (Chérif et al., 2023; Hasnain et al., 2022). Its flexibility and lower data requirements make it well-suited for low-resource settings.

Despite the growing use of both LSTM and Prophet, there is limited empirical comparison of their performance in African environmental contexts under severe data constraints. This gap restricts policymakers' ability to choose appropriate models for air quality management.

This study aims to address this knowledge gap by evaluating and comparing the predictive performance of LSTM and Prophet models across multiple pollutants and states in Northern Nigeria using observational data from 2018 to 2023. It seeks to determine the conditions under which each model performs best, hypothesising that Prophet will excel in forecasting pollutants dominated by seasonal and trend components, while LSTM will be better suited to capturing abrupt structural changes in pollutant variability. Ultimately, the research strives to provide actionable guidance for model selection in resource-constrained environmental monitoring contexts, promoting a data-driven approach that values robustness, simplicity, and adaptability over complexity. The findings intend to support more pragmatic and effective air quality management in developing regions.

## 2. Materials and methods

### 2.1 Study area

We selected the capital cities of 19 northern states of Nigeria, focusing on locations with available air quality data and sufficient meteorological data to ensure good spatial representativeness of results. These cities represent diverse geographic, demographic, and economic profiles, capturing variations in air pollution levels and their sources across the region. The selected cities range in size from medium to large urban centres, with populations varying significantly, and encompass distinct climatic conditions that influence air pollution dispersion. In these areas, the climate is typically tropical, characterised by seasonal variations, including a wet season (June to September) and a dry season (October to May), with annual temperatures ranging from 20°C to 40°C and rainfall varying from 600 mm to 1,200 mm depending on the location.

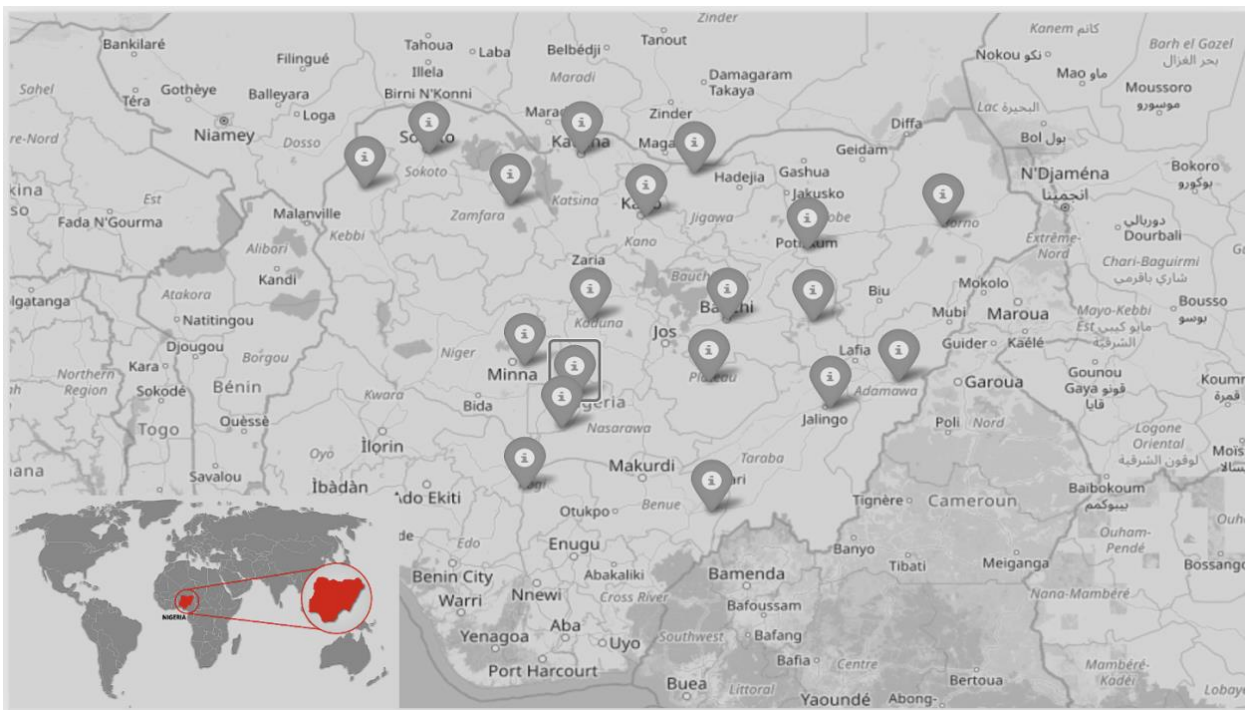

Figure 1: Location of selected cities in Northern Nigeria (created by author)

For instance, Kano, one of the most populous cities in Northern Nigeria and a major economic hub, has a population of over four million citizens within 449km². Its primary sources of pollution include vehicular emissions, industrial activities and burning waste. Maiduguri, the capital of Borno State, has a population of over eight hundred thousand and spans 105 km². The city's air pollution is largely driven using solid fuels for cooking and dust from unpaved roads. Similarly, Kaduna, located in the north-central region, is a significant industrial and administrative city with a population of over 1.2 million, covering an area of 431 km². Significant sources of air pollution in Kaduna include factory emissions, traffic-related pollution, and domestic energy use.

## 3. Methodology

This study focuses on predicting air pollutant concentrations across 19 states in Northern Nigeria using regression-based supervised learning. Supervised machine learning typically involves two iterative stages: training and testing, as illustrated in Figure 2.

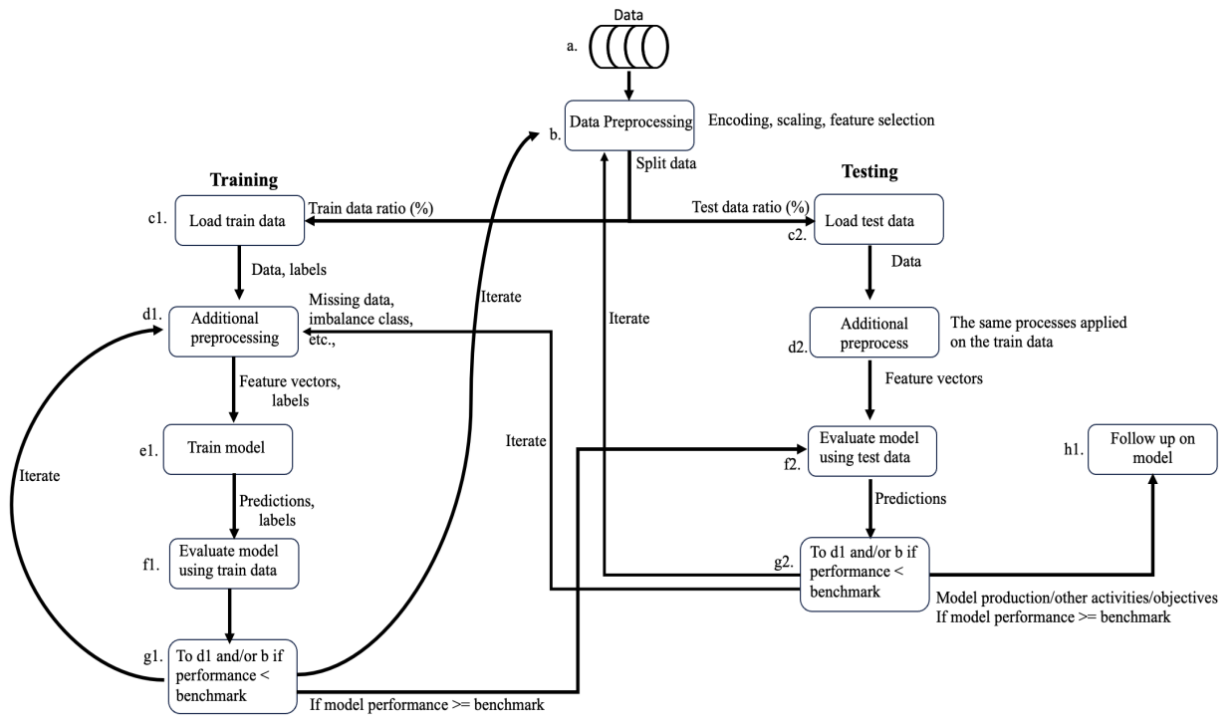

Figure 2: Supervised learning workflow (Balogun et al., 2025)

During the training phase, a model is built using labelled data by selecting relevant features, handling missing values, applying preprocessing steps (e.g., encoding, scaling), choosing a suitable algorithm, and tuning hyperparameters for optimal performance. The model's performance is initially assessed on the training data to detect underfitting or overfitting.

In the testing phase, the model is evaluated on an independent, unseen test set to gauge its generalizability and robustness. This evaluation helps ensure the model is not overly tailored to the training data and performs well on new inputs. If performance does not meet the benchmark, further iterations are conducted by revisiting preprocessing, feature selection, or model tuning.

As depicted in Figure 3 (Balogun et al., 2025), this workflow involves steps like data preprocessing, feature vector extraction, iterative model evaluation, and benchmarking. Models that meet performance criteria proceed to deployment or further objectives.

This study compares two models: Long Short-Term Memory (LSTM) networks and the Facebook Prophet Model (FPM). Both are assessed using real atmospheric pollutant measurements based on their predictive accuracy, robustness to data irregularities, and suitability for real-world forecasting tasks.

## 3.1 Data Collection and Feature Composition

### 3.1.1 Air quality data

Air quality data for this study were sourced from monitoring stations in the capital cities of the 19 northern states of Nigeria, using monthly averages from 2018 to 2023. The dataset included concentrations of $SO_2$, $CO$, $CO_2$, $SO_4$, PM2.5, and PM10 (all in µg/m³), along with meteorological variables such as wind speed (m/s), temperature (°C), and rainfall (mm). Monitoring stations covered 25–70 km² for particulate matter and 80–100 km² for gaseous pollutants. These data provide the foundation for analysing air quality trends and developing predictive models tailored to Northern Nigeria.

### 3.1.2 Meteorological data

Meteorological data for this study were collected for the 19 northern states of Nigeria, covering key variables such as wind speed, temperature, and rainfall. These data were sourced alongside air quality measurements from monitoring stations located in the capital cities of the 19 states, using monthly averages from 2018 to 2023. The meteorological dataset complements pollutant concentrations, enabling a comprehensive analysis of air quality trends and the development of predictive models tailored to Northern Nigeria. This combination of data supports understanding the relationships between meteorological factors and air pollutant levels.

## 3.2 Data Preprocessing

The dataset underwent thorough preprocessing to ensure model suitability. Missing values were handled using linear interpolation for continuous variables and forward-filling for seasonal lags. Features were scaled using Min-Max normalisation to aid convergence and prevent dominance. An 80/20 chronological train-test split simulated real-world forecasting scenarios.

## 3.3 Predictive Modelling Framework

### 3.3.1 Long Short-Term Memory (LSTM)

Long Short-Term Memory (LSTM) networks are a type of recurrent neural network (RNN) designed to capture sequential dependencies through gated memory units. Each LSTM cell retains both a cell state and a hidden state, allowing the network to preserve long-term contextual information across time steps. In this study, the LSTM architecture comprises a multivariate input layer incorporating pollutant and meteorological features, followed by one or two stacked LSTM layers with 64–128 memory cells. A dropout layer with a rate of 0.2 is included for regularisation to mitigate overfitting, and a fully connected dense layer produces the final pollutant concentration forecasts. The model is trained using the Adam optimiser with the mean squared error (MSE) as the loss function. Key hyperparameters, including learning rate, number of epochs, and batch size, are optimised through grid search based on validation loss minimisation.

### 3.3.2 Facebook Prophet Model (FPM)

The Facebook Prophet Model is an additive time series forecasting model expressed as:

$$y(t) = g(t) + s(t) + h(t) + \varepsilon_t$$

Where:
$g(t)$: piecewise linear or logistic growth trend
$s(t)$: seasonal component modelled using a Fourier series
$h(t)$: effects of holidays or known events
$\varepsilon_t$: random error term

Prophet automatically detects trend changepoints, handles missing data, and accommodates outliers through robust estimation. In this study, weekly and yearly seasonality were enabled, and no external regressors were added beyond pollutant observations.

### 3.4  Model Evaluation Metrics

The performance of each model was assessed using the following metrics:
Mean Squared Error (MSE): Measures average squared difference between observed and predicted values.
Root Mean Squared Error (RMSE): Square root of MSE; more interpretable due to original units.

$$RMSE = \sqrt{\frac{1}{n} \sum (y_i - \hat{y}_i)^2}$$

Mean Absolute Percentage Error (MAPE): Indicates average absolute per cent deviation from true values.

$$MAPE = \frac{1}{n} \sum \left( \left| (y_i - \hat{y}_i) / y_i \right| \right) \times 100$$

Coefficient of Determination ($R^2$): Captures the proportion of variance explained by the model.
Where $y_i$ is the actual value, $\hat{y}_i$ is the predicted value, and $n$ is the total number of observations.

### 3.5  Experimental Design

To ensure comparability, each model was trained on the same dataset and tasked with forecasting pollutant levels over an identical test window (2023–2024). Predictions were generated separately for each pollutant and location to capture spatial and chemical variability. Results were analysed both globally, across all 19 states, and locally, at the individual state level, to evaluate model consistency and generalisation.

# 4. Results and Discussion

## 4.1 Exploratory data analysis

Figure 3 presents summary statistics for the air pollutants across the 19 Northern states of Nigeria.

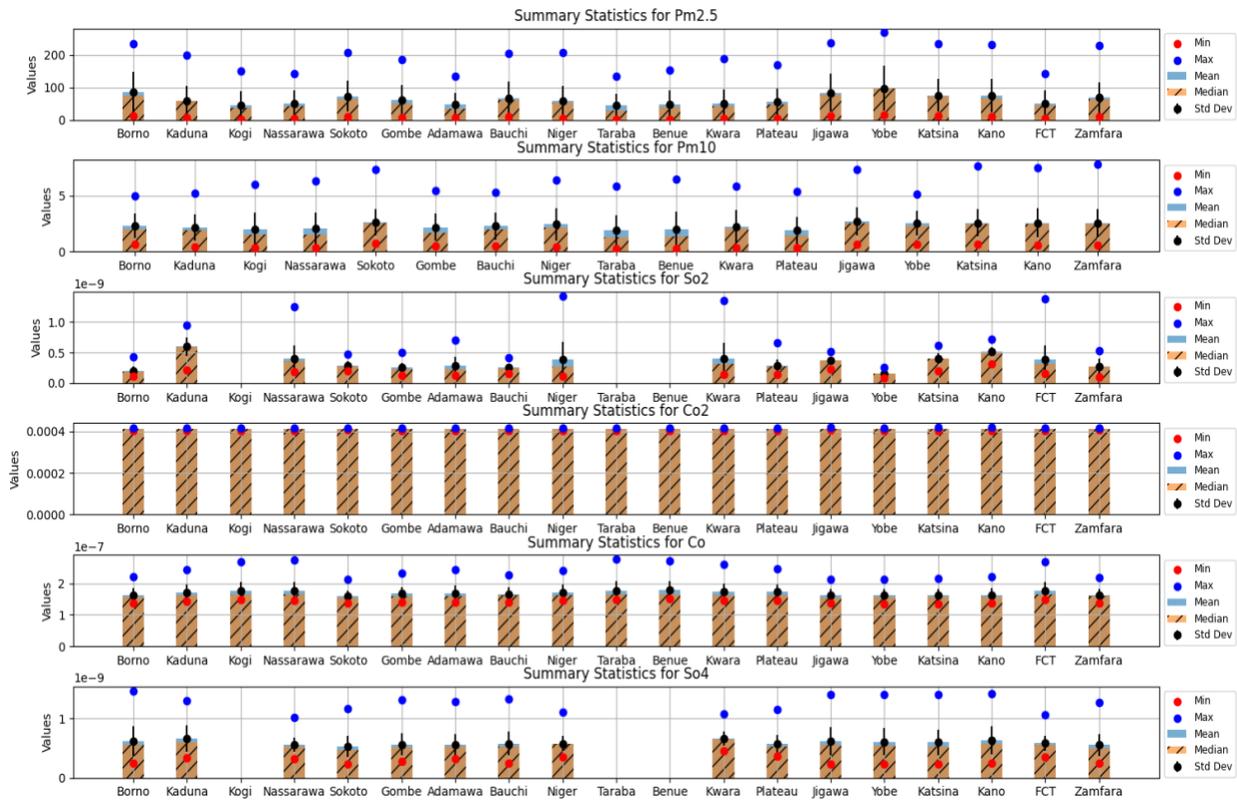

Figure 3. Summary Statistics of Air Pollutants Across the 19 Northern States

What really stands out is how varied the data quality and pollutant levels are from state to state. Some pollutants, like CO, $SO_2$, and $SO_4$, show very low or nearly constant values in many places, which suggests there might be issues with data collection or monitoring in those regions. On the other hand, other pollutants have more variation, but overall, the figure really highlights the challenge we face with inconsistent data across such a large geographic area.

Figure 4 indicates the temporal distribution of CO emissions across Northern Nigeria and shows potential cyclical pollution-level patterns. The cyclical nature of CO levels could be influenced by dry or wet seasons, agricultural practices (e.g., burning biomass), and variations in industrial activities. Large spikes might be associated with specific events like fires, industrial surges, or increased vehicular emissions during certain periods.

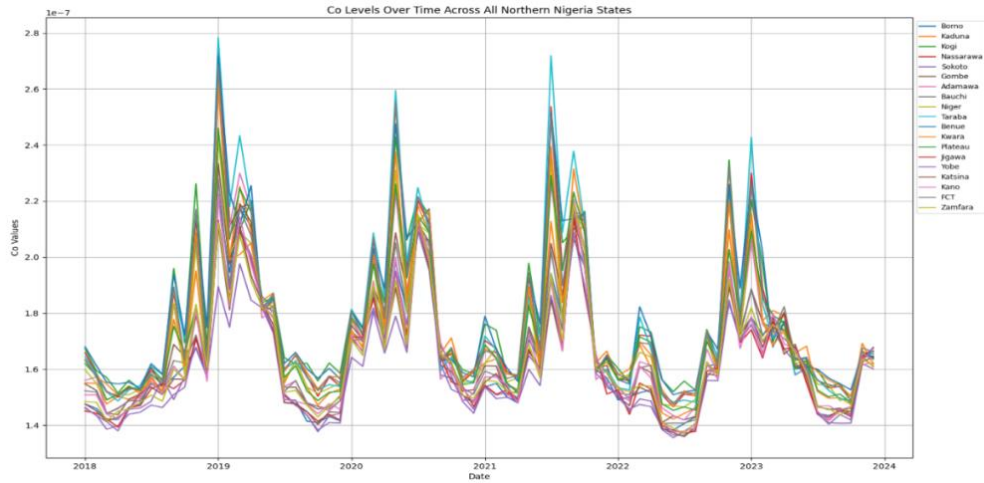

Figure 4: CO levels for all northern Nigeria states from 2018-2024 (source: Author)

Figure 5 is a time plot of the observed $CO_2$ level across the states. A noticeable upward trend in $CO_2$ levels is observed during the period with fluctuations, and the overall pollutant levels have been steadily increasing with subtle differences between the states, suggesting that $CO_2$ emissions are influenced by common regional factors across the states, potentially attributed to increased industrialization, population growth, or deforestation, which is very common in northern Nigeria.

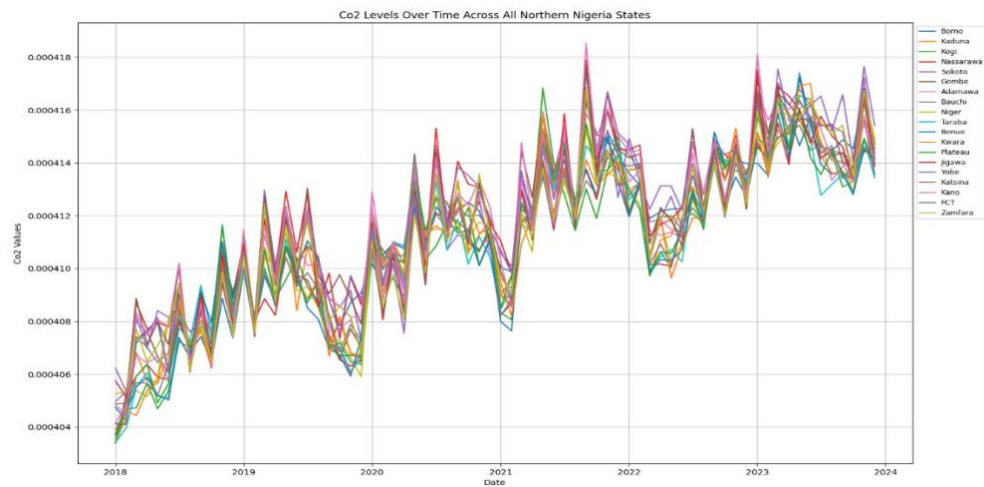

Figure 5: $CO_2$ levels for all northern Nigeria states from 2018-2024 (source: Author)

$PM_{10}$ time plot depicted in Figure 6 shows seasonal or cyclical fluctuations in PM10 levels across northern Nigerian states, with recurrent peaks likely driven by regional environmental factors such as dust storms during dusty trade wind or Harmattan season, that blows across the region from the Sahara, which could explain the regular spikes in PM10 levels during certain times of the year.

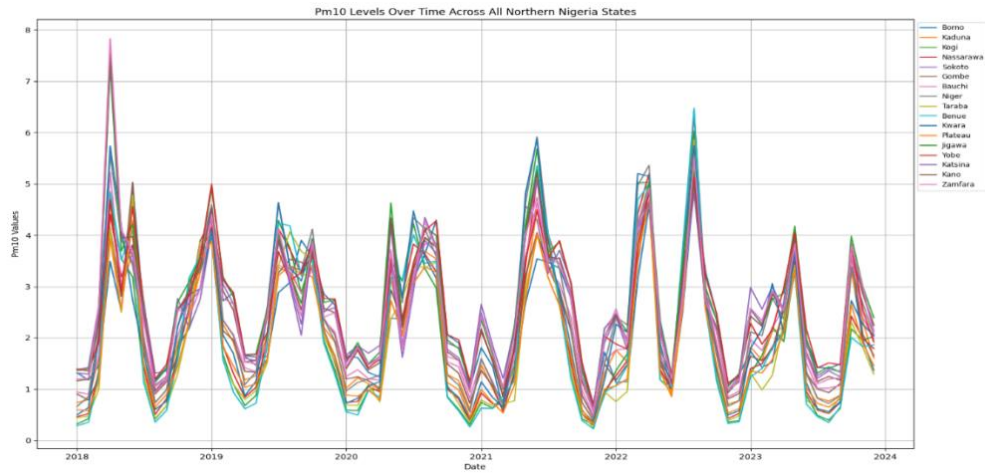

Figure 6: $PM_{10}$ levels for all northern Nigeria states from 2018-2024 (source: Author)

Different states have different concentrations of $SO_2$ levels, and higher concentrations were recorded between 2018 and early 2019, as can be seen from Figure 7 below. Seasonal and location environmental factors may potentially influence the variability of the concentration levels.

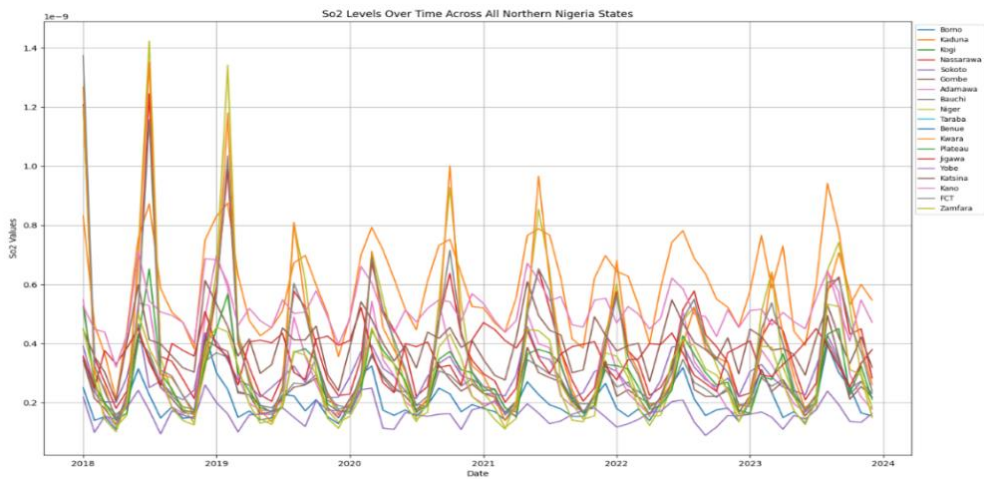

Figure 7: $SO_2$ levels for all northern Nigeria states from 2018-2024 (source: Author)

In Figure 8 below, dramatic fluctuations in $SO_4$ over most of the states and fewer deviations in the pattern are apparent. Common factors contributing to the pollutant concentration can be reasonably found in the northern states.

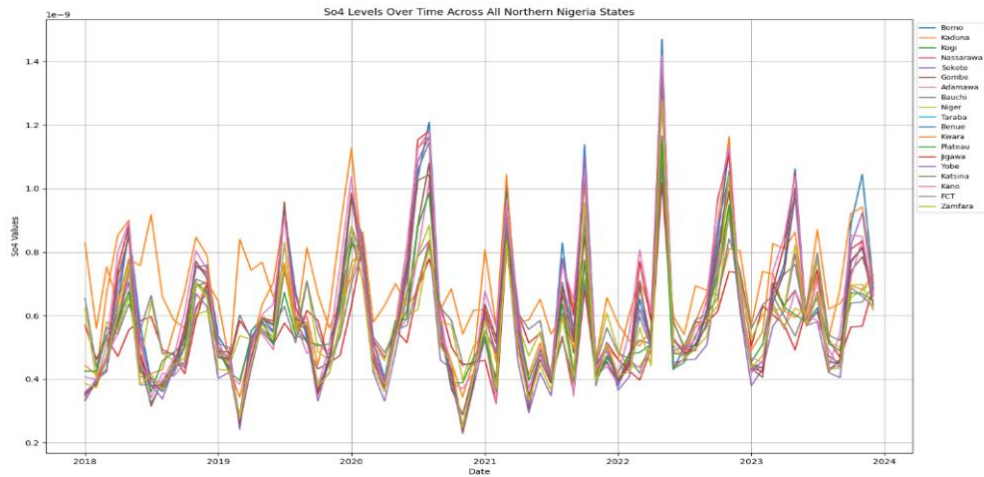

Figure 8: SO$_4$ levels for all northern Nigeria states from 2018-2024 (source: Author)

Following the exploratory data analysis, particularly Figure 3, this study specifically focuses on three key pollutants, CO, SO$_2$, and SO$_4$, in six selected states, despite the known challenges posed by their data quality. Although these pollutants exhibit extremely low values or near-constant readings across much of the dataset, indicating limited or potentially unreliable measurements, we deliberately chose to include them to showcase the capabilities of advanced forecasting models such as Facebook Prophet and LSTM.

Data limitations are widespread across the broader dataset covering 19 states in Northern Nigeria, with inconsistent or sparse pollutant records particularly evident for CO, SO$_2$, and SO$_4$. However, rather than excluding these pollutants due to their limited variability, we aimed to test how well these sophisticated time series models can learn from and predict patterns in challenging data environments.

By focusing on six states where data for these pollutants is relatively more available, albeit still exhibiting low variation, we provide a meaningful context to evaluate the robustness and adaptability of Facebook Prophet and LSTM. This approach highlights the power of these models to extract useful insights and generate forecasts even when data is noisy, sparse, or constrained, which is critical for real-world applications in resource-limited settings.

In summary, our targeted focus on CO, SO$_2$, and SO$_4$, combined with the selection of six states, strikes a balance between practical data challenges and the demonstration of model performance. This strategy reinforces the potential of advanced forecasting techniques to overcome data quality issues and deliver valuable predictions in contexts where pollutant measurements are often limited or unreliable. The next section presents the model predictions using LSTM and Facebook Prophet for the pollutants CO, SO$_2$, and SO$_4$ across six states: Kaduna, Sokoto, Jigawa, Katsina, Kano, and Zamfara.

## 4.2 Results: Model predictions
### 4.2.1 Carbon Monoxide (CO)

Observed CO (Figures 9 and 10) displays high-frequency peaks likely driven by traffic emissions and localised combustion. Both models captured these short-term fluctuations, though Prophet forecasts were marginally smoother.

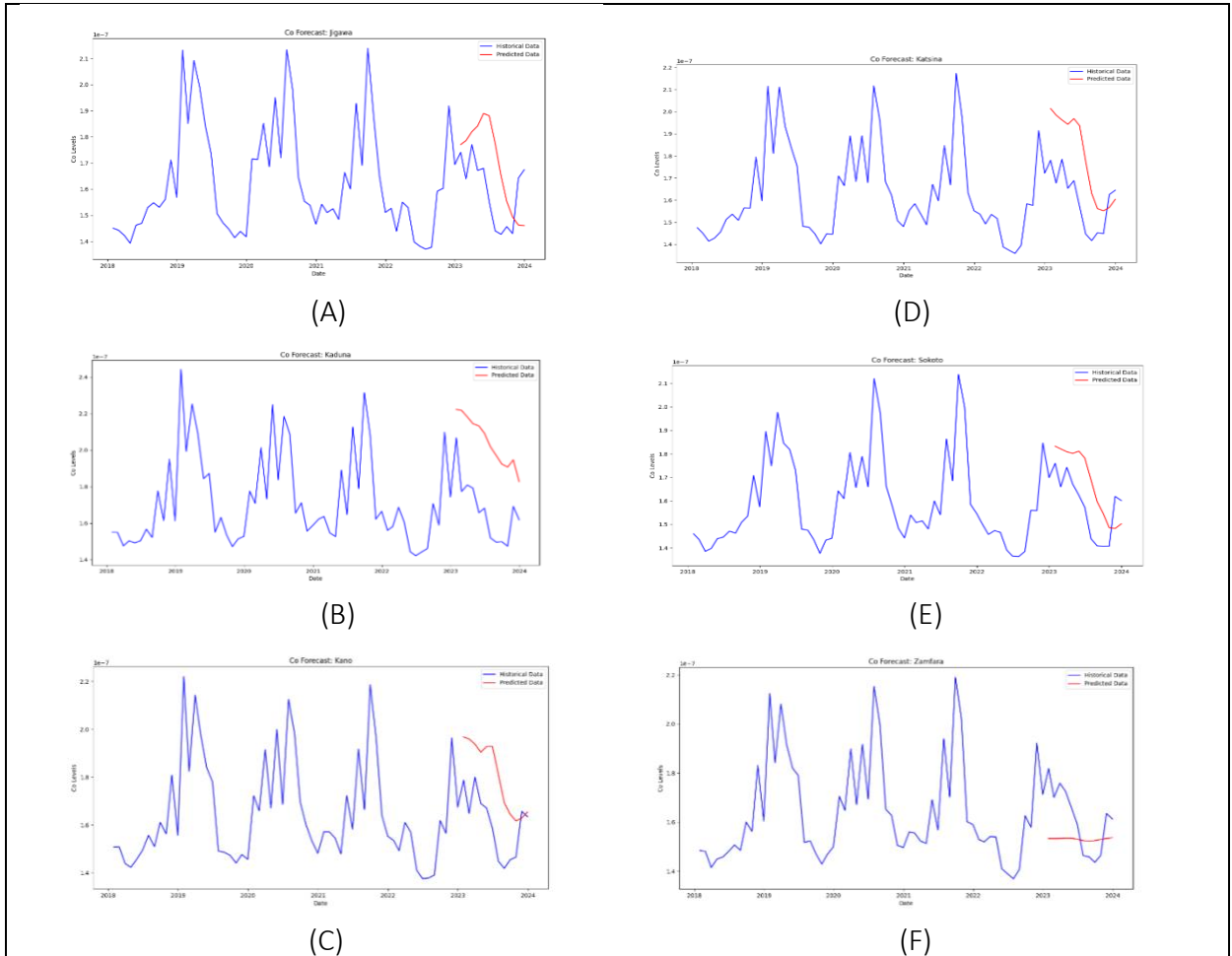

Figure 9: LSTM prediction for C0

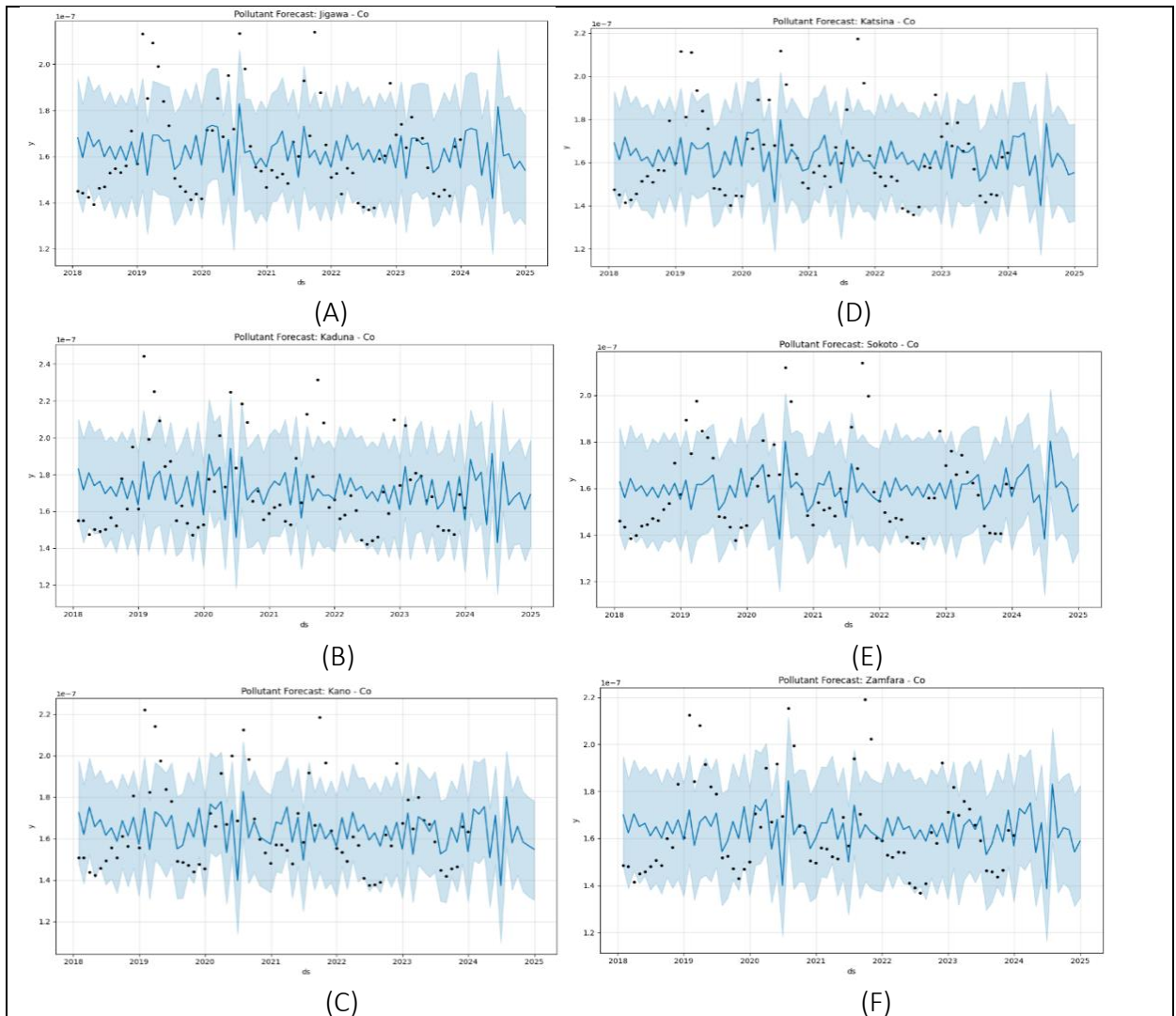

Figure 10: Prophet Prediction for C0

RMSE results (Table 1) indicate Prophet outperforms LSTM in Kaduna, with other states showing no statistically significant difference. LSTM exhibited slightly elevated RMSE in Sokoto, possibly due to data sparsity disrupting its temporal dependency learning.

Table 1: C0 RMSE VALUES

| RMSE FOR C0 POLLUTANT | | |
|---|---|---|
| STATE | LSTM | PROPHET |
| KADUNA | 3.82E-08 | 2.21E-08 |
| SOKOTO | 1.21E-08 | 1.73E-08 |
| JIGAWA | 1.53E-08 | 1.53E-08 |
| KATSINA | 2.07E-08 | 2.07E-08 |
| KANO | 2.14E-08 | 2.14E-08 |
| ZAMFARA | 1.53E-08 | 1.53E-08 |

### 4.2.2 Sulphur Dioxide (SO₂)

Historical SO₂ (Figures 11 and 12) reveal strong cyclical behaviour, likely from seasonal industrial output or meteorological dispersion patterns. LSTM forecasts suggest slight increases in 2024 for Jigawa, Kaduna, Kano, and Katsina, but stability in Sokoto and Zamfara. Prophet's forecasts largely replicate these dynamics but with smoother phase transitions.

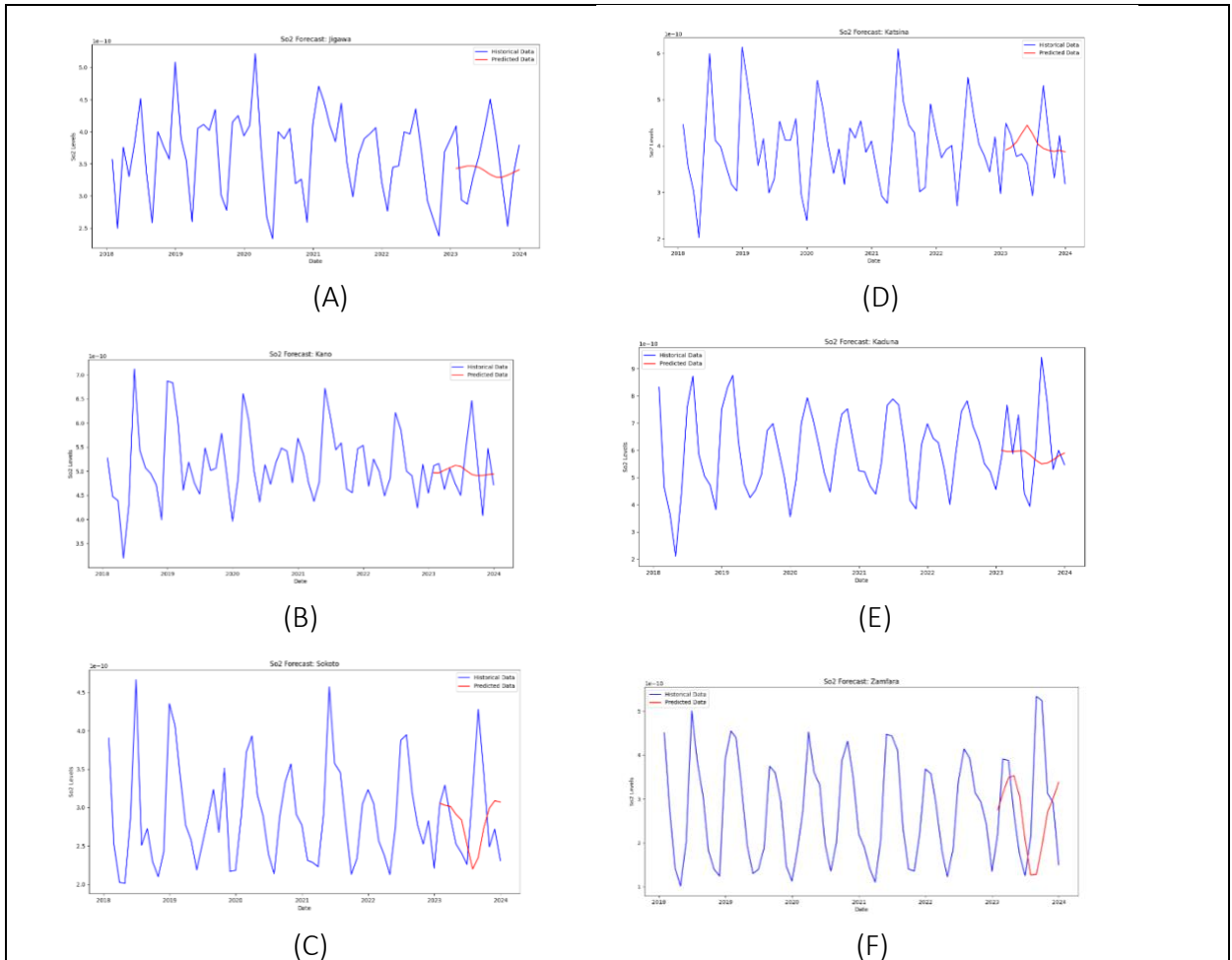

Figure 11: LSTM prediction for S0$_2$.

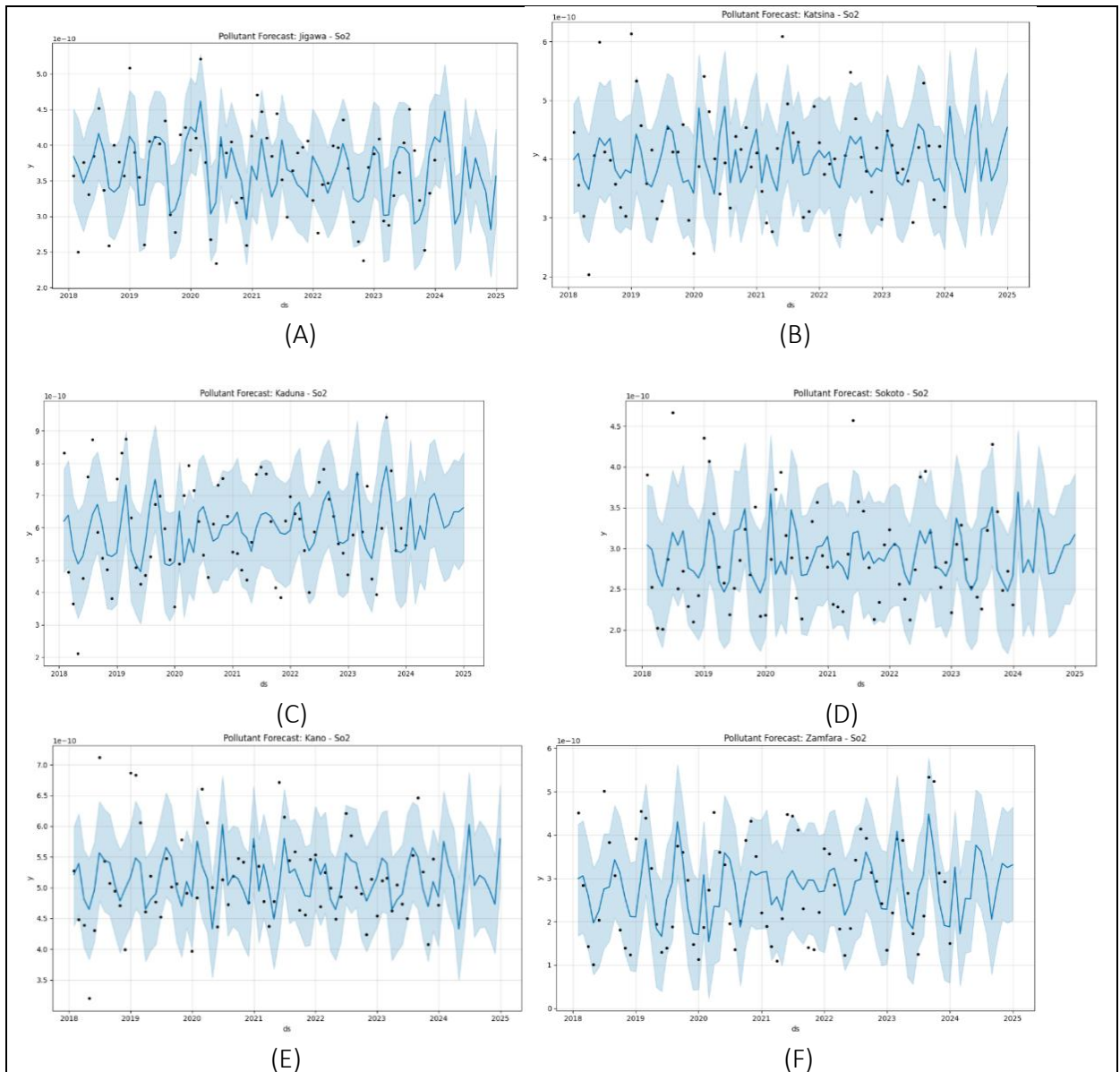

Figure 12: Prophet Prediction for $SO_2$.

Table 2 shows both models performing poorly in Kaduna and Zamfara (high RMSE), possibly reflecting unmodelled emission shocks. In other states, performance was strong, with Prophet slightly outperforming LSTM in cyclical but relatively low-noise series.

Table 2: $SO_2$ RMSE VALUES

| RMSE FOR SO2 POLLUTANT | | |
|---|---|---|
| STATE | LSTM | PROPHET |
| KADUNA | 1.63E-10 | 1.28E-10 |
| SOKOTO | 6.30E-11 | 5.80E-11 |
| JIGAWA | 6.06E-11 | 6.06E-11 |
| KATSINA | 7.10E-11 | 7.10E-11 |

| KANO | 6.28E-11 | 6.28E-11 |
| ZAMFARA | 1.37E-10 | 1.37E-10 |

### 4.2.3 Sulphate (SO₄)

SO₄ concentrations (Figures 14 and 14) dropped sharply between 2020 and 2021, likely reflecting pandemic-related emission reductions, followed by sharp rebounds in 2022 and 2023. Both models forecast continued declines with reduced variability.

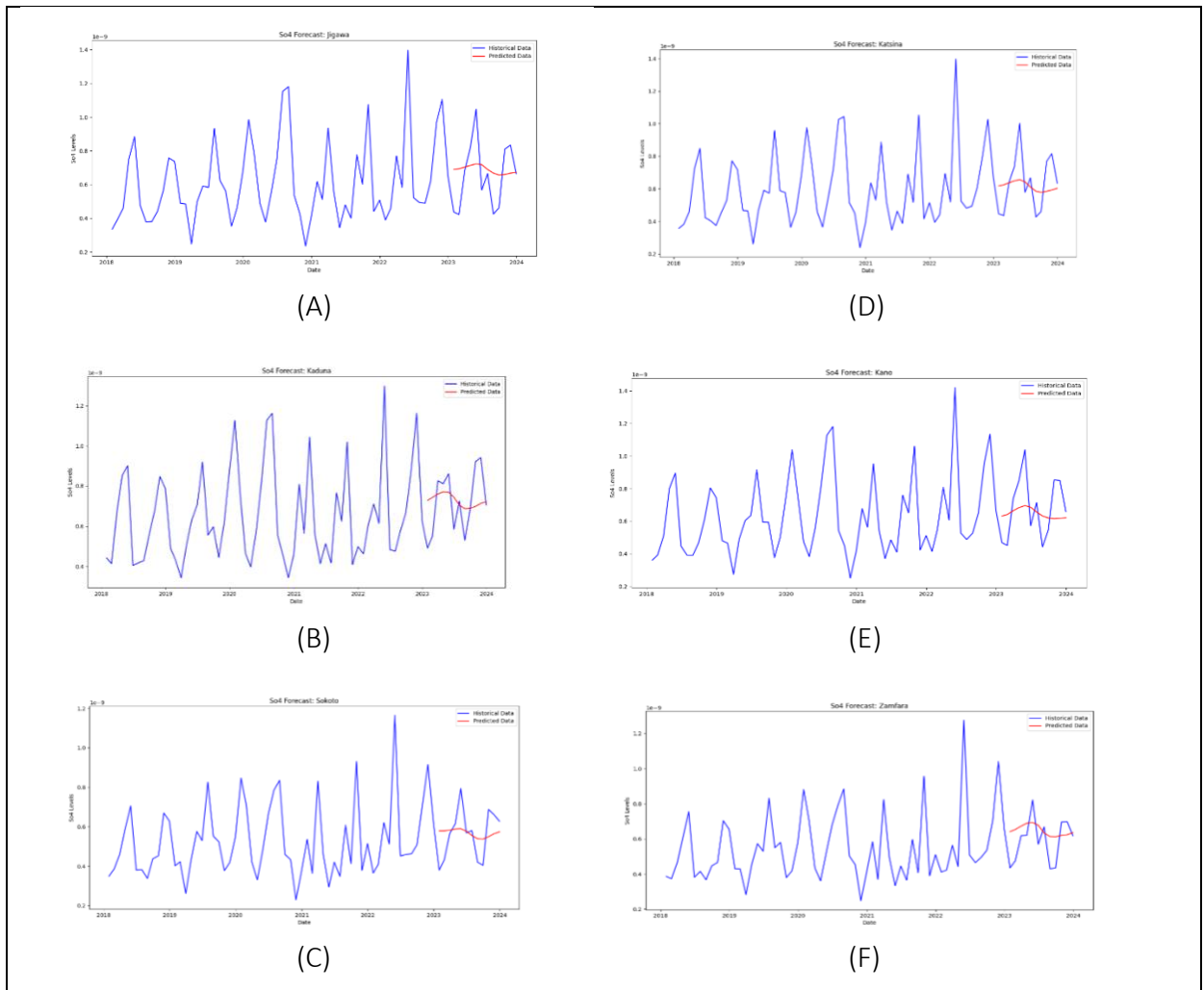

(A)　(D)
(B)　(E)
(C)　(F)

Fig. 4.23: LSTM prediction for S0₄.

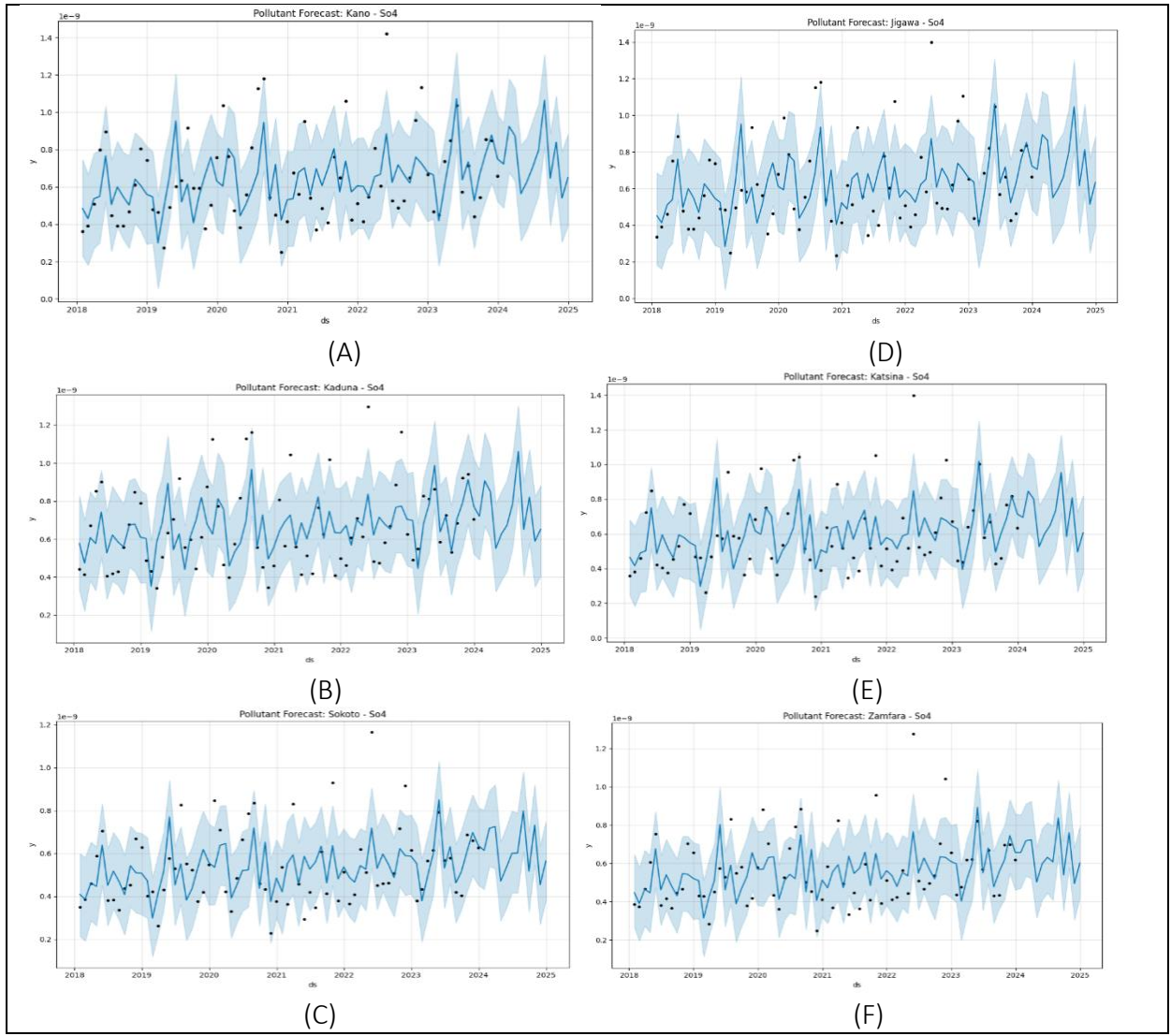

Figure 14: Prophet Prediction for $SO_4$

RMSE results (Table 3) show LSTM outperforming Prophet in Kaduna and Sokoto, with equal performance elsewhere. This suggests that LSTM's ability to learn from abrupt level shifts gave it a relative edge in pollutants exhibiting structural breaks.

Table 3: $SO_2$ RMSE VALUES

| RMSE FOR SO4 POLLUTANT | | |
|---|---|---|
| STATE | LSTM | PROPHET |
| KADUNA | 1.50E-10 | 1.88E-10 |
| SOKOTO | 1.19E-10 | 1.45E-10 |
| JIGAWA | 1.87E-10 | 1.87E-10 |
| KATSINA | 1.67E-10 | 1.67E-10 |
| KANO | 1.79E-10 | 1.79E-10 |
| ZAMFARA | 1.30E-10 | 1.30E-10 |

## 4.3 Discussion

This study critically examines the relative performance of two prominent forecasting models, Facebook Prophet and Long Short-Term Memory (LSTM) networks, in predicting air pollutant concentrations across 19 states in Northern Nigeria, a region typified by data scarcity and resource constraints. While the dataset covers a broad spatial domain, this analysis specifically focuses on three key pollutants: carbon monoxide (CO), sulphur dioxide ($SO_2$), and sulphate ($SO_4$) within six selected states (Kaduna, Sokoto, Jigawa, Katsina, Kano, and Zamfara). These pollutants were chosen due to their importance in regional air quality, despite exhibiting low or near-constant values and data irregularities across much of the dataset. The six states were selected based on relatively higher data unavailability for these pollutants, providing a practical yet challenging context to evaluate model performance.

The findings challenge the widespread presumption that complex deep learning models invariably outperform simpler, statistically grounded approaches in environmental time series forecasting. Prophet's consistent performance, often matching or surpassing LSTM, particularly for pollutants exhibiting strong seasonal patterns and long-term trends, corroborates previous research emphasising its robustness in the face of irregular data and missing observations (Chérif et al., 2023; Hasnain et al., 2022). This resilience stems from Prophet's decomposable time series structure, which naturally models trend and seasonality components, making it particularly suited for operational contexts where data completeness and quality are often compromised.

Conversely, LSTM demonstrated superior adaptability in scenarios involving abrupt structural shifts, such as observed for $SO_4$ in Kaduna and Sokoto, validating its strength in capturing complex temporal dependencies and regime changes (Peralta et al., 2022). However, LSTM's higher model complexity demands substantial data quantity and quality, as well as rigorous hyperparameter tuning, which may limit its practical application in low-resource environments, consistent with cautionary notes raised by Jin et al. (2021).

From a practical perspective, the computational efficiency and interpretability of Prophet provide compelling advantages for environmental agencies operating under resource constraints. Its minimal preprocessing requirements and tolerance for missing data simplify deployment, enabling more immediate and cost-effective forecasting solutions. Nonetheless, when short-term pollutant fluctuations and sudden environmental shocks are the forecasting focus, the LSTM model's capacity for nuanced temporal modelling justifies its additional complexity, provided adequate expertise and data availability.

## 4.4 Conclusion

This comparative analysis underscores the necessity of context-specific model selection in air pollution forecasting within resource-constrained settings. By focusing on CO, $SO_2$, and $SO_4$ across six Northern Nigerian states selected for relatively low data availability amidst widespread measurement challenges, the study highlights Prophet's competitive performance across diverse urban environments. Prophet's strength in forecasting pollutants dominated by seasonal and trend components affirms its utility as a pragmatic and efficient forecasting tool. Meanwhile, LSTM's strengths in handling abrupt regime shifts highlight its value where detailed temporal dynamics and rapid fluctuations are central concerns.

Strengths of this study include its broad geographic coverage, the use of real-world datasets characterised by irregularities typical of low-resource contexts, and the methodological contrast between statistical and deep learning models.

However, limitations arise from the use of monthly averaged data, potentially obscuring critical short-term pollution spikes. Additionally, the absence of hybrid or ensemble approaches limits the exploration of combined model benefits. Future work should focus on leveraging higher-resolution temporal data (daily or hourly) to better capture short-term variability, integrating exogenous variables such as meteorological factors and satellite-derived metrics, and developing hybrid models that combine Prophet's interpretability with LSTM's flexibility. Expanding the geographic scope to other African regions and performing cost-benefit analyses of model deployment would further enhance generalizability and operational relevance.

## Acknowledgement

The authors would like to thank Muhammad Zayyanu for his invaluable assistance in obtaining and providing access to the air pollution data for the 19 Northern Nigeria states used in this study.